\documentclass[journal]{IEEEtran}
\usepackage{cite}
 \usepackage{array}
\usepackage{amsmath,amssymb,amsfonts}
\usepackage{algorithmic}
\usepackage{comment}

\usepackage{makecell}    
\usepackage{overpic}
\usepackage{comment}
\usepackage{paralist}
\usepackage{verbatim}

\usepackage{bm} % for bold math like \bm{\theta}

\usepackage{float}        % For [H] specifier to force table position
\usepackage{graphicx}     % For any image/table scaling (commonly needed)
\usepackage{booktabs}   % For \toprule, \midrule, \bottomrule
\usepackage{multirow}   % For multirow cells in    % For pseudocode

\usepackage{cite}              % For compact IEEE-style citations
\usepackage{comment}
\usepackage[ruled,vlined,linesnumbered]{algorithm2e}
\SetKwInput{KwInput}{Input}
\SetKwInput{KwOutput}{Output}

\usepackage{tikz}
\usetikzlibrary{arrows.meta,positioning,shapes.geometric,fit}
\usepackage{pgfplots}
\pgfplotsset{compat=1.18}
\usepackage{subcaption}

 \usepackage{enumitem}      % for compact enumerate/itemize

\def\Htran{\mbox{\tiny $\mathrm{H}$}}
\def\Ttran{\mbox{\tiny $\mathrm{T}$}}
\def\BibTeX{{\rm B\kern-.05em{\sc i\kern-.025em b}\kern-.08em
    T\kern-.1667em\lower.7ex\hbox{E}\kern-.125emX}}

\begin{document}
%\title{ Evolutionary Optimization Frameworks for  Near-Field Multi-Source Localization}
\title{ A Primer on Evolutionary Optimization Frameworks for Near-Field Multi-Source Localization}

\author{
Seyed Jalaleddin Mousavirad, \IEEEmembership{Senior Member, IEEE};
Parisa Ramezani, \IEEEmembership{Member, IEEE};
 Mattias O'Nils, \IEEEmembership{Member, IEEE}; and
 Emil Björnson, \IEEEmembership{Fellow, IEEE}
\thanks{ Seyed Jalaleddin Mousavirad and Mattias O'Nils are with Department of Computer and Electrical Engineering, Mid Sweden University, Sundsvall, Sweden (e-mail: \{seyedjalaleddin.mousavirad, mattias.onils\}@miun.se). Parisa Ramezani is with the school of electrical and data engineering, University of Technology Sydney, Sydney, Australia (email: parisa.ramezani@uts.edu.au). Emil Bj\"ornson is with the Department of Computer Science, KTH Royal Institute of Technology, Stockholm, Sweden (e-mail:  emilbjo@kth.se).
This work was supported by the Swedish Knowledge Foundation and by the SUCCESS project (FUS21-0026), funded by the Swedish Foundation for Strategic Research.} }

\maketitle

\begingroup
\renewcommand\thefootnote{}
\footnotetext{This work has been submitted to the IEEE for possible publication. Copyright may be transferred without notice, after which this version may no longer be accessible.}
\addtocounter{footnote}{-1}
\endgroup

\begin{abstract}

This paper introduces evolutionary optimization as a grid-free training-free continuous-domain search mechanism for near-field multi-source localization, addressing the major limitations of grid-based subspace methods such as MUSIC and data-driven deep learning approaches. To this end, we develop two complementary evolutionary localization frameworks that operate directly on the continuous spherical-wave signal model and support arbitrary array geometries without requiring labeled data, discretized angle-range grids, or architectural constraints. The first framework, termed NEar-field MultimOdal DE (NEMO-DE) associates each individual in the evolutionary population to a single source and optimizes a residual least-squares objective in a sequential manner, updating the data residual and enforcing spatial separation to estimate multiple source locations. To overcome the limitation of NEMO-DE under large power imbalances among the sources, we propose the second framework, named NEar-field Eigen-subspace Fitting DE (NEEF-DE), which jointly encodes all source locations and minimizes a subspace-fitting criterion that aligns a model-based array response subspace with the received signal subspace. The proposed formulations are not intrinsically tied to a specific optimizer; however, this work adopts differential evolution (DE) as a representative evolutionary search strategy because of its simple implementation, small number of control parameters, and strong empirical performance in continuous nonconvex optimization problems. Numerical results show that the proposed frameworks provide competitive accuracy compared with MUSIC-type baselines while avoiding pre-defined grid construction and labeled training data. This work establishes evolutionary computation as a powerful and flexible paradigm for model-based near-field localization, paving the way for future innovations in this domain.

\end{abstract}


\section{Introduction}
\label{sec:introduction}
Localization is a key enabler for many wireless applications and use cases, including industrial automation \cite{Liang2018Hybrid,Batalla2020Adaptive}, health monitoring \cite{Eid2021Holography}, emergency rescue and public-safety operations \cite{Zhang2024LoRaAid}, and a wide range of other location-based services where reliable position information is essential.

Among established approaches are time-of-arrival/time-difference-of-arrival (ToA/TDoA) \cite{Kang2023An,Sadeghi2021Optimal}, their enhanced versions using ultra-wideband (UWB)-based signaling \cite{Martalo2022Improved,Verde2023Characterization}, and fingerprint-based localization \cite{Tian2017Performance,Zhou2021Exploiting,Xie2024Environment}. Despite their strengths, each of these methods possesses important drawbacks. TDoA demands tightly synchronized anchors, and ToA requires source-anchor synchronization or uses two-way ranging, which adds extra message exchanges to cancel clock bias.
While UWB signaling substantially improves time resolution, it does not change synchronization requirements. On the other hand, the accuracy of fingerprinting typically depends on labor-intensive site surveys and degrades in case of environmental changes. 

The recent move toward larger antenna arrays has opened new doors for precise localization by expanding the so-called radiative near-field region and bringing more sources into a regime where spherical-wave models enable joint angle-range estimation \cite{Wang2025Near}. This near-field capability can reduce reliance on tight network time synchronization and dense, site-wide fingerprint maintenance, addressing several limitations of ToA/TDoA and fingerprinting.

MUltiple SIgnal Classification (MUSIC) is a high-resolution subspace-based method widely used for localization because it can resolve and localize multiple sources simultaneously. Originally developed for angle-of-arrival (AoA) estimation  \cite{Schmidt1986Multiple}, it has been extended to near-field localization with a uniform linear array (ULA) by adopting a spherical-wave model in which the array response vector depends on both angle and range \cite{Huang1991Near}. The near-field MUSIC forms a two-dimensional (2D) pseudospectrum over angle and range, and identifies the peaks to localize multiple sources. This requires evaluating the pseudospectrum on a discrete angle-range grid, creating a trade-off between computational burden and the error introduced by grid mismatch.
Several techniques have been proposed to reduce the grid search complexity of MUSIC. Reference \cite{Zhi2007Near} splits a ULA into symmetric subarrays and uses a generalized Estimation of Signal Parameters via Rotational Invariance Technique (ESPRIT) to get all angles in a one-dimensional (1D) search step, followed by a per-source 1D MUSIC over range. 
The authors in \cite{He2012Efficient} exploit the array symmetry and the anti-diagonal structure of the covariance matrix to decouple angle and range estimation problems. They first construct a vector that only depends on AoAs and utilize 1D MUSIC to estimate the angles. Substituting the estimated angles back into the original 2D MUSIC pseudospectrum, source ranges are obtained via independent 1D searches. 
A reduced-dimension MUSIC is proposed in \cite{Zhang2018Localization}, which performs 1D angle searches and then computes paired ranges in closed form. The work in \cite{Zuo2019Localization} aims to alleviate the computational burden associated with search-based approaches by proposing a new linear prediction method. It estimates auxiliary phase parameters via linear prediction applied to the cross-anti-diagonal covariance terms, recovers angle and range from the zeros of a prediction polynomial, and then refines the estimation with an oblique-projection iterative scheme. Although these works reduce the search burden of MUSIC, they rely on specific array structures, e.g., less than a quarter of wavelength spacing between the antennas, limiting their practical applicability. 

Another active line of work employs deep learning to replace or augment classical searches for near-field localization. The work in \cite{Lee2022Deep} proposes a deep learning-based method for near-field source localization, which consists of two main components: a Principal Component Analysis Network (PCAnet) and a Spatial Spectrum Network (Sp2net). The PCAnet computes noise spaces from the received signal covariance matrix using convolutional layers, while Sp2net generates a high-resolution spatial spectrum using a linear layer and a custom activation function that enhances values at source locations while being differentiable. A convolutional neural network (CNN) then classifies dense azimuth and range grids to estimate source angles and ranges. \cite{Gast2025DCD} introduces a deep learning-aided cascaded differentiable MUSIC algorithm for near-field localization of multiple sources, which combines deep neural networks with classical subspace methods. The method uses two deep neural networks; one to generate a surrogate far-field covariance for estimating the number of sources and their angles via ESPRIT, and another to produce a covariance for range estimation via 1D MUSIC.
A CNN-based approach is presented in \cite{Ramezani2025Machine} for three-dimensional (3D) near-field source localization with a uniform planar array (UPA). Their proposed method directly maps the eigenvectors of the received signal's covariance matrix to the sources' 3D coordinates, thereby avoiding the computationally expensive 3D grid search required by the conventional MUSIC algorithm by learning a continuous mapping from the input data.
While these deep learning-based near-field localization methods achieve high accuracy and efficiency, they face generalization limits: their performance is contingent on the test conditions matching the training setup, as they learn directly from a specific labeled dataset and may struggle with unseen scenarios.
%\textcolor{red}{Although deep learning–based near-field localization methods can offer high accuracy and computational efficiency, they suffer from fundamental generalization limitations. Their performance is heavily tied to the specific conditions represented in the training data, making them brittle when faced with variations in array geometry, propagation environments, source configurations, or noise statistics. As a result, these models often fail to maintain accuracy in unseen scenarios, require costly labeled datasets, and lack the robustness needed for reliable deployment in real-world industrial environments.}

%\textcolor{red}{To handle the limitations of deep learning models and to develop a training-free, fully model-driven approach, this paper proposes two evolutionary localization frameworks} 
%Jalal: Here we can talk about evolutionary
These limitations motivate the search for a training-free, model-driven localization framework that operates directly on the near-field array response and avoids discretized angle-range grids. In this paper, the term grid-free means that the proposed methods do not require a predefined  angle-range grid. Instead, the unknown source parameters are optimized directly over the continuous feasible domain. This does not eliminate numerical search; rather, it replaces exhaustive evaluation over a fixed grid with an adaptive population-based search that concentrates the computational effort on promising regions of the parameter space. As a result, the proposed framework mitigates grid-mismatch effects and can provide a more flexible accuracy–complexity tradeoff, particularly in high-dimensional near-field localization scenarios.

The near-field localization problem is well suited to this formulation because the spherical-wave signal model gives rise to a coupled angle-range parameter space with highly nonconvex and multimodal objective landscapes. Multiple coexisting sources naturally correspond to multiple candidate optima in this continuous domain. Based on this observation, we develop two complementary evolutionary localization frameworks. The first framework, NEMO-DE, treats near-field multi-source localization as a sequential multimodal optimization problem: each individual encodes one candidate source location, and multiple sources are detected as distinct optima through residual fitting, projection-based deflation, and distance-based penalization. The second framework, NEEF-DE, treats the problem as a joint global optimization problem: each individual encodes the complete multi-source configuration, and all source locations are estimated simultaneously by minimizing a subspace-fitting criterion. Thus, the main contribution is not the selection of a particular evolutionary optimizer, but the design of problem-specific representations, objective functions, and search mechanisms that adapt evolutionary optimization to grid-free near-field multi-source localization. Although different evolutionary or population-based optimizers can be incorporated into the proposed frameworks, this paper adopts differential evolution (DE)~\cite{storn1995differential,price2006differential} as a representative search strategy due to its simplicity, robustness, and reliable performance in nonconvex continuous optimization problems.

Our main contributions are summarized as follows.
\begin{itemize}
    
    \item \textbf{Model-driven grid-free formulation:} We formulate near-field multi-source localization as a model-driven evolutionary optimization problem over the continuous spherical-wave parameter space. In contrast to grid-based MUSIC-type searches, the proposed formulation does not require a predefined  angle-range grid. In contrast to data-driven approaches, it does not require labeled training data and operates directly on the physical near-field array response.
       \item \textbf{Sequential multimodal residual fitting:} We develop a compact-representation scheme in which each individual encodes a single source, and solve the near-field localization problem via sequential evolutionary searches that minimize a data-domain residual reconstruction error in the least-squares (LS) sense. After each detection, a projection-based deflation updates the residual, and a distance-based penalization prevents duplicate modes, yielding efficient discovery of multiple sources.
    \item \textbf{Joint eigen-subspace fitting.} We develop NEEF-DE, an expanded-representation evolutionary framework in which each individual jointly encodes the complete multi-source configuration. The localization task is treated as a joint global optimization problem by minimizing a subspace-misalignment criterion between the model-based near-field array response subspace and the received signal subspace. 
    \end{itemize}
%To the authors’ best knowledge, this work is the first attempt to systematically develop an evolutionary computation (EC)-based framework for wireless near-field multi-source localization, and to design both sequential and joint evolutionary search strategies for estimating the sources’ continuous location parameters.

%To the best of the authors' knowledge, evolutionary optimization has received limited attention as a continuous-domain search mechanism for wireless near-field multi-source localization. The objective functions used in this work are inspired by classical residual-decomposition and subspace-fitting principles, including CLEAN \cite{hogbom1974clean}, Matching Pursuit \cite{mallat1993matching}, WSF \cite{viberg1991wsf}, and MODE-type formulations \cite{stoica1990mode}. However, they are adapted here to a different near-field evolutionary localization setting. Specifically, NEMO-DE performs sequential residual fitting over the continuous spherical-wave manifold using DE, while incorporating a distance-based penalty to reduce duplicate detections. NEEF-DE, in contrast, performs joint subspace fitting by optimizing all source locations simultaneously in the continuous near-field parameter space. Thus, the main contribution is the adaptation of these classical estimation ideas into two grid-free evolutionary frameworks for near-field multi-source localization, highlighting their different accuracy--complexity--robustness tradeoffs.

The remainder of this paper is organized as follows: Section~\ref{sec:EC} reviews EC and one of its representative algorithms, DE.  Section~\ref{sec:sysmod} presents the system model and problem setup. The proposed DE-based near-field localization frameworks are introduced in Section~\ref{sec:NEMO-DE} and Section~\ref{sec:NEEF-DE}, respectively.
The computational complexity analysis of the proposed schemes is presented in Section~\ref{sec:complexity}.  
Section~\ref{sec:evaluation} provides a comprehensive numerical evaluation, comparing the proposed schemes with well-known near-field localization algorithms. 
Finally, Section~\ref{sec:conc} concludes this paper.

\section{Evolutionary Computation}
\label{sec:EC}
EC represents a family of population-based stochastic optimization techniques inspired by the principles of natural evolution and genetics. These algorithms operate through mechanisms analogous to biological evolution, such as selection, mutation, recombination, and inheritance, to evolve candidate solutions toward optimal or near-optimal results. EC methods have been widely employed in complex optimization problems where analytical solutions are difficult or infeasible to obtain, particularly in multimodal, non-differentiable, or noisy environments~\cite{mousavirad2025robust, mousavirad2024evolutionary}.

%jalal: algorithm agnostic
%The proposed localization frameworks are \emph{algorithm-agnostic}, as they are defined by problem-specific solution representations and objective functions rather than by the choice of a particular evolutionary solver. Consequently, they can be integrated with a wide range of evolutionary or population-based optimization methods. 
In this study, DE~\cite{storn1995differential} is employed as a representative algorithm due to its conceptual simplicity, strong global search capability, and consistently competitive performance across diverse optimization domains, although other evolutionary algorithms could also be used. 
%making it a widely used benchmark for evaluating EC-based methodologies.

%The proposed approach is algorithm-agnostic, enabling its integration with any existing EC framework. However, in this study, DE~\cite{storn1995differential} is employed as a representative algorithm. DE is chosen due to its conceptual simplicity, strong global search capability, and consistently competitive performance across diverse optimization domains, making it a widely recognized benchmark for evaluating new EC-based methodologies.

\subsection{Differential Evolution}
\label{sec:DE}
DE is a simple yet powerful population-based optimization algorithm \cite{storn1995differential}, particularly effective for continuous parameter optimization. The original DE evolves a population of candidate solutions through three key operations: \textit{mutation}, \textit{crossover}, and \textit{selection}. 

 In the standard variant of DE, DE/rand/1/bin, at each generation $g$, a mutant vector is generated by adding the weighted difference between two population members to a third one, as follows:
\begin{equation}
\mathbf{v}_{i}^{(g)} = \mathbf{x}_{r_1}^{(g)} + F \times (\mathbf{x}_{r_2}^{(g)} - \mathbf{x}_{r_3}^{(g)}),
\end{equation}
where $\mathbf{v}_{i}^{(g)}$ denotes the mutant vector for the $i$-th individual at generation $g$, $F \in [0,2]$ is the mutation scaling factor, and $r_1, r_2, r_3$ are distinct indices randomly selected from the population.

Next, a \textit{crossover} operation combines the target vector $\mathbf{x}_{i}^{(g)}$ and its corresponding mutant vector $\mathbf{v}_{i}^{(g)}$ to form a trial vector $\mathbf{u}_{i}^{(g)}$. The binomial (uniform) crossover is defined as:
\begin{equation}
u_{i,j}^{(g)} =
\begin{cases}
v_{i,j}^{(g)}, & \text{if } \text{rand}_j(0,1) \leq C_r \text{ or } j = j_{\text{rand}}, \\
x_{i,j}^{(g)}, & \text{otherwise},
\end{cases}
\end{equation}
where $C_r \in [0,1]$ is the crossover probability, $j$ denotes the index of the decision variables (i.e., the dimension index) within the vector, $\text{rand}_j(0,1)$ is a uniformly distributed random number for the $j$-th dimension, and $j_{\text{rand}}$ is a randomly selected dimension index that ensures at least one parameter is inherited from the mutant vector.

Finally, \textit{selection} determines which vector survives to the next generation. A greedy selection mechanism ensures that only the better vector (in terms of the objective function) is retained:
\begin{equation}
\mathbf{x}_{i}^{(g+1)} =
\begin{cases}
\mathbf{u}_{i}^{(g)}, & \text{if } f(\mathbf{u}_{i}^{(g)}) \leq f(\mathbf{x}_{i}^{(g)}), \\
\mathbf{x}_{i}^{(g)}, & \text{otherwise},
\end{cases}
\end{equation}
where $f(\cdot)$ denotes the objective (fitness) function to be minimized.

Through these iterative steps, DE efficiently balances \textit{exploration} (via mutation and crossover) and \textit{exploitation} (via selection), leading to robust performance across diverse optimization landscapes.

%jalal we did not talk about glovbal optimisation! 
\subsection{Global and Multimodal Optimization}

Optimization is the process of determining decision variables that minimize or maximize an objective function over a feasible search space. In most classical settings, the goal is to identify a single best solution, referred to as a global optimum. Such problems are known as global optimization problems. Classical optimization techniques, including many evolutionary algorithms, are therefore designed to guide the search process toward this dominant solution.

A global optimization problem can be formulated as
\begin{equation}
\min_{x \in \Omega} f(x),
\end{equation}
where $f: \Omega \subseteq \mathbb{R}^n \rightarrow \mathbb{R}$ is the objective function and $\Omega$ denotes the feasible search space. A point $x^* \in \Omega$ is called a global minimizer if
\begin{equation}
f(x^*) \leq f(x), \quad \forall x \in \Omega.
\end{equation}
For maximization problems, the inequality is reversed accordingly.

However, in some problems, not only the global optimum is important, but local optima are also of significant interest. In such cases, identifying only the single best solution is insufficient. These problems are referred to as MMO problems~\cite{preuss2015multimodal, liu2018survey}. Unlike global optimization, MMO aims to locate and preserve multiple optima, both local and global, that are distributed across different regions of the search space~\cite{wong2015evolutionary}.

Formally, a function $f$ is said to be multimodal if it possesses multiple local minima within $\Omega$. A point $x_i^* \in \Omega$ is called a local minimizer if there exists a neighborhood $\mathcal{N}(x_i^*)$ such that
\begin{equation}
f(x_i^*) \leq f(x), \quad \forall x \in \mathcal{N}(x_i^*).
\end{equation}

The objective of MMO is therefore to identify a set of optimal solutions
\begin{equation}
\mathcal{S}^* = \{ x_1^*, x_2^*, \dots, x_k^* \},
\end{equation}
where each $x_i^*$ represents a distinct local or global optimum belonging to different attraction basins of the objective landscape.

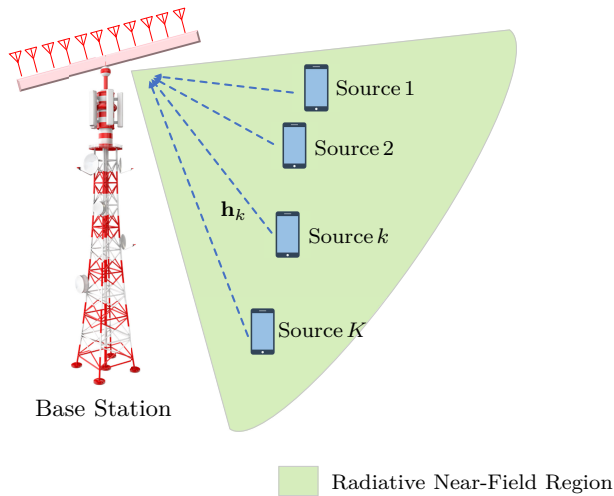
\begin{figure}[t!]
  \centering
   \begin{overpic}[scale = 0.13]{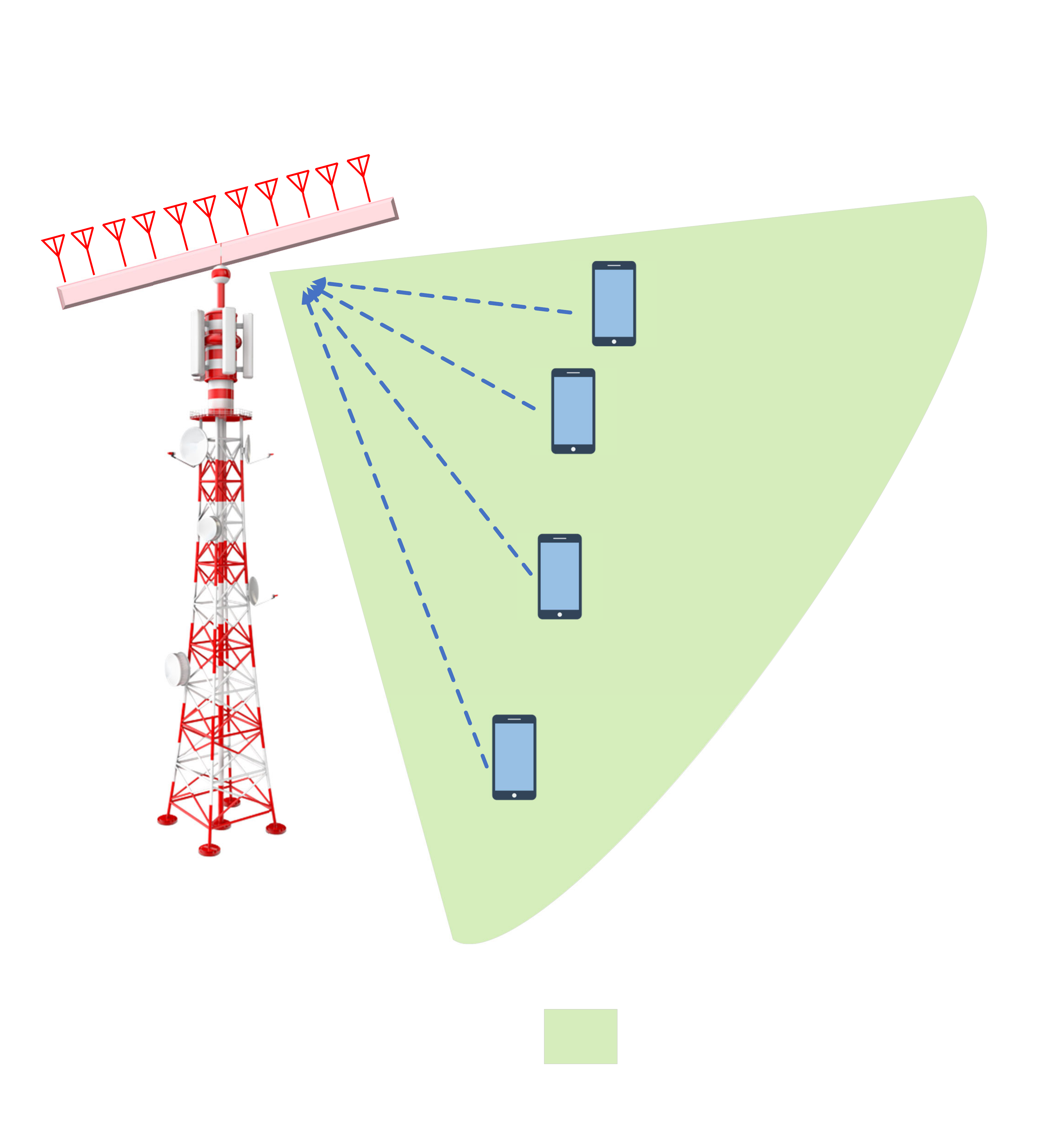}
  \put(8,19){\small Base Station}%
  \put(57.5,72){\footnotesize Source\,$1$}%
   \put(54,62){\footnotesize Source\,$2$}%
   \put(53,47.5){\footnotesize Source\,$k$}%
  \put(48,32){\footnotesize Source\,$K$}%
  \put(56.8,7.1){\footnotesize Radiative Near-Field Region}%
  \put(38.5,52){\footnotesize $\mathbf{h}_k$}
   \end{overpic}
\caption{System model for near-field  multi-source localization.}
\label{fig:system}
\vspace{-3mm}
\end{figure}

 \section{System Model}
 \label{sec:sysmod}
 We consider the scenario depicted in Fig.~\ref{fig:system}, where $K$ sources, located in the radiative near-field region of a base station (BS) with $M$ antennas in the form of a ULA, transmit  independent signals. The near-field boundary is characterized by the well-known Fraunhofer distance, computed as $d_{\mathrm{FA}} = 2D_{\mathrm{ap}}^2/\lambda$, where $D_{\mathrm{ap}}$ is the aperture length of the array and $\lambda$ is the carrier wavelength. The signals are captured by the BS, with the received signal at the BS at time slot $t$ being expressed as  
\begin{equation}
\label{eq:received_signal}
  \mathbf{y}(t) = \sum_{k=1}^K \mathbf{a}(\phi_k,r_k) s_k(t) + \mathbf{n}(t), 
\end{equation}
where $\mathbf{a}(\phi_k,r_k) \in \mathbb{C}^{M}$ is the near-field array response vector for the signal coming from source~$k$ with $\phi_k$ and $r_k$ being the azimuth AoA and range of the source, respectively, and $\mathbf{n}(t) \in \mathbb{C}^M$ is the additive independent complex Gaussian noise at the BS. The near-field array response vector is given by 
\begin{equation}
\label{eq:steering_vec}
\mathbf{a}(\boldsymbol{\theta}) = 
\exp\!\left(j\frac{2\pi}{\lambda}(d_1(\boldsymbol{\theta}) - \mathbf{d}(\boldsymbol{\theta}))\right),
\end{equation}
where $\boldsymbol{\theta} = [\phi,r]^{\Ttran}$ is the location parameter vector, $d_1(\boldsymbol{\theta})$ is the distance between the candidate source and the reference BS antenna and \(\mathbf{d}(\boldsymbol{\theta}) = [d_1(\boldsymbol{\theta}),\ldots,d_M(\boldsymbol{\theta})]^{\Ttran}\) contains the distances from the source to all $M$ BS antennas, with 
\begin{equation}
   d_m(\boldsymbol{\theta}) \approx  r - (m-1)\delta \sin(\phi) + \frac{(m-1)^2 \delta^2}{2r}, 
\end{equation}
where $\delta$ denotes the spacing between array antennas. 
Assuming that the sources transmit over $T$ time slots, the collected signal at the BS is
\begin{equation}
\label{eq:collected_signal}
\mathbf{Y} = \mathbf{A}\mathbf{S} + \mathbf{N},     
\end{equation}
where $\mathbf{Y} = [\mathbf{y}(1),\ldots,\mathbf{y}(T)] \in \mathbb{C}^{M \times T}$, $\mathbf{A} = [\mathbf{a}(\boldsymbol{\theta}_1),\ldots,\mathbf{a}(\boldsymbol{\theta}_K)] \in \mathbb{C}^{M\times K}$, $\mathbf{S} = [\mathbf{s}_1,\ldots,\mathbf{s}_K]^{\Ttran} \in \mathbb{C}^{K\times T}$ with $\mathbf{s}_k = [s_k(1),\ldots,s_k(T)]^{\Ttran}$, and $\mathbf{N} = [\mathbf{n}(1),\ldots,\mathbf{n}(T)] \in \mathbb{C}^{M \times T}$. 

In \eqref{eq:received_signal}, we assume a pure line-of-sight (LoS) scenario where the channels are only characterized by the near-field array response vectors. Although all the subsequent analyses are based on this assumption, we move beyond the idealized pure LoS assumption in Section~\ref{sec:evaluation} when evaluating the performance of the proposed schemes, and adopt a more realistic Rician fading scenario where the LoS component remains dominant but is accompanied by a weaker non-LoS (NLoS) component. 
This is a typical scenario in localization problems, where a dominant LoS component enables reliable extraction of source location information from the BS near-field array response vectors.  Furthermore, the number of sources can be estimated using standard information-theoretic criteria for model selection such as Akaike Information Criterion (AIC) and Minimum Description Length (MDL) \cite{Wax1985Detection}. In this work, the number of active sources $K$ is assumed to be known when running the localization algorithms. This assumption is common in localization studies~\cite{stoica1990maximum}, where source enumeration is treated as a separate model-order selection problem. 

To this end, we introduce two fundamentally different evolutionary-based estimators for near-field multi-source localization: a sequential multimodal evolutionary strategy that detects sources as distinct local optima of a residual objective, and a joint global evolutionary strategy that simultaneously estimates all source locations by minimizing a multimodal subspace-fitting criterion.
%We propose two DE-based estimators for near-field multi-source localization: {\color{blue}a multimodal DE and a single-solution global DE (with a multimodal objective function).} 
It is worth mentioning that the proposed schemes readily extend to the more general case of a UPA-formed BS. In that setting, the location parameter vector becomes three-dimensional, i.e., $\boldsymbol{\theta} = [\phi,\psi,r]^{\Ttran}$, where $\psi$ denotes the elevation angle of the corresponding source.  

\section{Near-Field  Multimodal DE Localization}
\label{sec:NEMO-DE}
The first proposed DE framework for near-field multi-source localization, referred to as \textit{NEar-field MultimOdal DE (NEMO-DE)}, is designed to exploit the multimodal structure of the localization problem. 
NEMO-DE treats each source location as a distinct local optimum in the parameter space and applies DE as a population-based optimizer to discover multiple modes sequentially. 
%This approach leverages the inherent diversity-preserving capability of DE and integrates it with a residual-based objective that directly operates on the received data at the BS. 
%Unlike traditional grid-search or subspace-scanning methods, EvoLoc-MM explores the spatial domain adaptively, reducing computational redundancy while maintaining robustness to noise and initialization.

\subsection{Key Components}
The proposed NEMO-DE framework consists of three key components: the \textit{representation scheme} that defines how individuals encode candidate solutions, the \textit{objective function} that quantifies the data-domain consistency between the observed and modeled signals, and the \textit{multimodal evolutionary search} to find multiple distinct source locations. The main components of the proposed framework are explained in detail below. 

\subsubsection{Compact Representation}
In NEMO-DE, each individual in the evolutionary population encodes the spatial parameters of a single source.  
To localize multiple sources, NEMO-DE performs iterative evolutionary searches, where the residual signal after each detection is used as input for the next search. 
Such a sequential multimodal search mechanism allows the algorithm to successively uncover distinct local optima corresponding to different sources without requiring explicit clustering or mode assignment.

\subsubsection{Residual Least-Squares  Objective Function}

The objective function in NEMO-DE is formulated as a residual least-squares (RLS) criterion, which quantifies the mismatch between the received BS data and the reconstructed signal obtained from a candidate source hypothesis. This formulation is inspired by residual-decomposition methods such as CLEAN~\cite{hogbom1974clean} and Matching Pursuit~\cite{mallat1993matching}. However, NEMO-DE differs from conventional dictionary-based implementations in that its dictionary is not a fixed discrete set of atoms; instead, it is the continuous near-field spherical-wave manifold $\{\mathbf a(\phi,r)\}$, over which DE performs the off-grid search.

Given \(\mathbf{Y}\) in \eqref{eq:collected_signal}, the LS estimate of \(\mathbf{s}_k\) is
\begin{equation}
\hat{\mathbf{s}}_k^{\Ttran} = \left(\mathbf{a}^{\Htran}(\boldsymbol{\theta}_k) \mathbf{a}(\boldsymbol{\theta}_k)\right)^{-1}\mathbf{a}(\boldsymbol{\theta}_k)^{\Htran}\mathbf{Y}.
\end{equation}

The RLS objective is defined as the squared Frobenius norm of the residual between the received signal matrix and its LS reconstruction for a candidate $\boldsymbol{\theta}$:
\begin{equation}
\label{eq:RLS}
J_{\mathrm{RLS}}(\boldsymbol{\theta}) = 
\left\| (\mathbf{I} - \mathbf{P}_{\mathbf{a}}(\boldsymbol{\theta}))\mathbf{Y} \right\|_F^2,
\end{equation}
where $\mathbf{P}_{\mathbf{a}}(\boldsymbol{\theta}) = \mathbf{a}(\boldsymbol{\theta})\left(\mathbf{a}^{\Htran}(\boldsymbol{\theta}) \mathbf{a}(\boldsymbol{\theta})\right)^{-1}\mathbf{a}(\boldsymbol{\theta})^{\Htran}$ projects the received signal onto the subspace spanned by the candidate array response vector.
% toward the location of the estimated source. 
Minimizing \(J_{\mathrm{RLS}}(\boldsymbol{\theta})\) ensures that the estimated parameters produce the smallest reconstruction error between the received and modeled signals. 
Because the RLS objective function is typically highly nonconvex and exhibits multiple local minima corresponding to distinct sources, it provides a natural landscape for multimodal evolutionary search. 
Thus, each minimum of \(J_{\mathrm{RLS}}(\boldsymbol{\theta})\) corresponds to a potential source location in the near-field region of the BS.

%\subsection{Evo-Multi-SubSpace or Evo-Loc-E?}

\subsubsection{Multimodal Evolutionary Search}

The NEMO-DE algorithm employs a multimodal evolutionary strategy to identify multiple distinct source locations corresponding to different local minima of the RLS objective function. 
\subsection{Multimodal Evolutionary Process}
Let the search space be denoted by 
\(\boldsymbol{\Theta} \subseteq \mathbb{R}^2\),
where each parameter vector 
\(\boldsymbol{\theta}\) represents a candidate source location.
The optimization problem is formulated as
\begin{equation}
\boldsymbol{\theta}^\ast =
\arg\min_{\boldsymbol{\theta} \in \boldsymbol{\Theta}}
J_{\mathrm{RLS}}(\boldsymbol{\theta}).
\end{equation}

Although this formulation defines a single global minimizer,
the objective function \(J_{\mathrm{RLS}}(\boldsymbol{\theta})\) is generally
multimodal and exhibits multiple distinct local minima.
In the context of EC, which operates on a population of
solutions, this multimodal landscape gives rise to a set of locally optimal
candidate solutions
\(\{\boldsymbol{\theta}_1^\ast, \boldsymbol{\theta}_2^\ast, \ldots,
\boldsymbol{\theta}_K^\ast\}\),
each corresponding to a potential source location.

To explore this multimodal landscape, NEMO-DE employs DE as its optimization backbone. 
Each DE run evolves a population of candidate solutions over several generations through mutation, crossover, and selection, 
as previously described in Section~\ref{sec:DE}. 
The best-performing individual in each run, corresponding to the smallest residual cost, 
is identified as a detected mode \(\boldsymbol{\theta}_k^\ast\).

After each mode is detected, its contribution is subtracted from the received signal to allow the next search to focus on unexplored regions. 
This is achieved using a projection-based residual update:
\begin{equation}
\mathbf{Y} \leftarrow 
(\mathbf{I} - \mathbf{P}_{\mathbf{a}}(\boldsymbol{\theta}_k^\ast))\, \mathbf{Y}.
\end{equation}

To prevent redundant detections near previously identified sources, 
a distance-based penalization term is introduced in the fitness evaluation. 
Let 
\(\mathcal{S}_{\mathrm{det}} = \{\boldsymbol{\theta}_1^\ast, \ldots, \boldsymbol{\theta}_{k-1}^\ast\}\)
denote the set of already detected modes. 
The penalized objective for a new candidate \(\boldsymbol{\theta}\) is:
\begin{equation}
\tilde{J}_{\mathrm{RLS}}(\boldsymbol{\theta}) = 
J_{\mathrm{RLS}}(\boldsymbol{\theta}) +
\sum_{\boldsymbol{\theta}_i^\ast \in \mathcal{S}_{\mathrm{det}}}
\alpha \, \max(0, \delta_{\min} - \|\boldsymbol{\theta} - \boldsymbol{\theta}_i^\ast\|_2),
\end{equation}
where \(\delta_{\min}\) defines the minimum allowable separation between detected sources, 
and \(\alpha\) is a penalty coefficient controlling the repulsion strength. 
This penalization dynamically modifies the fitness landscape, guiding the DE search toward unexplored regions of the parameter space. Note that the penalization term $\|\boldsymbol{\theta} - \boldsymbol{\theta}_i^\ast\|_2$ is a normalized Euclidean norm, i.e., $\|\boldsymbol{\theta} - \boldsymbol{\theta}_i^\ast\|_2 = \|[(\phi - \phi_i^\ast)/\phi_0, (r - r_i^\ast)/r_0 ]\|_2$ with $\phi_0 = 1\,$rad and $r_0 = 1\,$m, so the resulting norm is dimensionless.

The multimodal evolutionary process proceeds iteratively. 
After each DE run, the best solution \(\boldsymbol{\theta}_k^\ast\) is stored, $\mathbf{Y}$ is updated using the residual projection, and a new search begins. 
The process continues until a predefined stopping condition is reached, 
such as a maximum number of detected sources.
%\begin{equation}
%\|\mathbf{Y}\|_F^2 < \varepsilon_{\text{stop}}.
%\end{equation}
Through this iterative search-update-penalize mechanism, 
NEMO-DE systematically uncovers all significant modes of the RLS landscape, 
thereby localizing multiple near-field sources without requiring prior information or supervised training.

Algorithm~\ref{Alg:NEMO} summarizes the proposed NEMO-DE localization framework.

\begin{algorithm}[t]
\caption{Multimodal RLS-based  DE Localization }
\label{Alg:NEMO}
\KwInput{Received data $\mathbf{Y}$, domain $\boldsymbol{\Theta}$, 
 population $N_p$, max generations $G_{\max}$, DE params $(F,C_r)$, 
 min-separation $\delta_{\min}$, penalty $\alpha$, number of sources $K$}
\KwOutput{Detected locations $\mathcal{S}_{\mathrm{det}} = \{\hat{\boldsymbol{\theta}}_1,\dots,\hat{\boldsymbol{\theta}}_K\}$}
\BlankLine
$\mathcal{S}_{\mathrm{det}} \leftarrow \emptyset$, $k\leftarrow 1$\;
\While{$k \le K$ }{
  \tcp{Define single-source residual objective (with penalization)}
  \For{candidate $\boldsymbol{\theta}$}{
    compute $J_{\mathrm{RLS}}(\boldsymbol{\theta};\mathbf{Y}) = \min_{\mathbf{s}}\|\mathbf{Y}-\mathbf{a}(\boldsymbol{\theta})\mathbf{s}^{\Ttran}\|_2^2$\;
    compute penalty $p(\boldsymbol{\theta}) = \sum_{\boldsymbol{\theta}'\in\mathcal{S}_{\mathrm{det}}}
      \alpha\max(0,\delta_{\min}-\|\boldsymbol{\theta}-\boldsymbol{\theta}'\|_2)$\;
    $\tilde{J}(\boldsymbol{\theta}) \leftarrow J_{\mathrm{RLS}}(\boldsymbol{\theta};\mathbf{Y}) + p(\boldsymbol{\theta})$\;
  }
  \tcp{Run DE (or any evolutionary optimizer) to minimize $\tilde{J}$ over $\boldsymbol{\Theta}$}
  Run DE with population $N_p$, generations $G_{\max}$ to obtain best $\hat{\boldsymbol{\theta}}_k$ and optionally refine by local search\;
  \tcp{Accept or reject detection }
  If detection quality is poor (e.g., $\tilde{J}(\hat{\boldsymbol{\theta}}_k)$ is large) then \textbf{break}\;
  \tcp{Update residual}
  $\hat{\mathbf{s}}_k^{\Ttran} \leftarrow \arg\min_{\mathbf{s}}\|\mathbf{Y}-\mathbf{a}(\hat{\boldsymbol{\theta}}_k)\mathbf{s}^{\Ttran}\|_2^2$\;
  $\mathbf{Y} \leftarrow \mathbf{Y} - \mathbf{a}(\hat{\boldsymbol{\theta}}_k)\hat{\mathbf{s}}_k^{\Ttran}$\;
  $\mathcal{S}_{\mathrm{det}} \leftarrow \mathcal{S}_{\mathrm{det}} \cup \{\hat{\boldsymbol{\theta}}_k\}$\;
  $k \leftarrow k+1$\;
}
\Return $\mathcal{S}_{\mathrm{det}}$
\end{algorithm}

%Even with a very dense grid (Grid 3), the proposed NEMO-DE still outperforms MUSIC, while requiring more than $197\times$ less runtime. The proposed scheme maintains high localization accuracy at a substantially lower computational complexity.

\section{Near-Field Eigen-Subspace Fitting DE Localization}
\label{sec:NEEF-DE}
In the previous section, we proposed a novel DE-based near-field localization method with an objective function defined as the residual between the received signal matrix and its LS reconstruction. By introducing a compact representation, the search strategy was formulated as an MMO problem.
In this section, we take a different approach by formulating the near-field multi-source localization as a subspace alignment problem.

The second DE framework, referred to as \textit{NEar-field Eigen-subspace Fitting DE (NEEF-DE)}, is designed to jointly estimate all sources' locations by exploiting the signal subspace structure of the received data. 
Unlike NEMO-DE, which performs sequential multimodal searches based on the data-domain residual, NEEF-DE adopts a global optimization strategy that aligns the model-based and data-derived signal subspaces in a single evolutionary process. 
%This formulation enables NEEF-DE to capture the spatial correlation among multiple sources while avoiding the need for iterative deflation or residual updates.

The motivation for proposing the NEEF-DE framework is the limited robustness of NEMO-DE in scenarios where the received SNRs of different sources are highly unbalanced. NEMO-DE performs sequential estimation by minimizing an energy-weighted RLS objective for one source at a time and then updating the data via projection/deflation before searching for the next source. Under pronounced SNR disparities, this sequential residual fitting becomes dominated by the strongest source(s) and even small modeling or estimation mismatches for a high-SNR source can leave non-negligible leakage in the residual, which distorts the cost landscape seen by subsequent searches. Meanwhile, the projection-based residual update can remove part of the weaker sources’ components along with the detected strong source, reducing their effective energy in the residual. As a result, later searches may fail to expose clear modes associated with weak sources, leading to missed detections and error propagation across iterations. This sensitivity of NEMO-DE to SNR imbalance is evaluated in Section~\ref{sec:evaluation}.
To mitigate this sensitivity, NEEF-DE adopts a joint evolutionary search over all sources in a single optimization, thereby avoiding sequential deflation and reducing dependence on relative source powers.

\begin{comment}
The motivation for proposing a subspace fitting multi-source localization framework stems from the limitations of NEMO-DE in handling scenarios with significant SNR variations among the sources. 
To illustrate this, we consider a ULA-formed BS of $M = 128$ antennas with $\lambda/2$ inter-antenna spacing at the carrier frequency of $15\,$GHz, and $K = 3$ sources where one has a fixed SNR of $15\,$dB, and the other two have SNRs of $15 - \delta_s\,$dB and $15+\delta_s\,$dB. Fig.~\ref{fig:rmse_snrdev} plots RMSE of NEMO-DE as a function of SNR deviation $\delta_s$. The results show that the RMSE increases rapidly with $\delta_s$, demonstrating that the localization accuracy is highly sensitive to SNR variations among the sources.
The reason is that NEMO-DE  fits sources sequentially using an energy-weighted RLS objective; small mismatches on strong sources leave leakage in the residual, and the projection step also removes part of the weak sources’ energy. Therefore, weak sources become harder to recover.
\begin{figure}
    \centering
    \includegraphics[width=0.95\linewidth]{rmse_snrdev.eps}
    \caption{RMSE of the NEMO-DE scheme under unequal received SNRs across sources.}
\label{fig:rmse_snrdev}
\vspace{-3mm}
\end{figure}
\end{comment}

%To alleviate the above issue, we propose a subspace fitting DE that jointly estimates all source locations by minimizing the mismatch between the received signal subspace and the model subspace spanned by the candidate array response matrix, making the search power-invariant and avoiding sequential deflation. 
\subsection{Key Components}
Similar to NEMO-DE, NEEF-DE consists of three main components: the \textit{representation scheme}, the \textit{objective function}, and the \textit{evolutionary search process}. However, their definitions are fundamentally different. 

\subsubsection{Expanded Representation}
In NEEF-DE, each individual in the evolutionary population represents the complete set of parameters corresponding to all $K$ sources. 
The individual vector is expressed as
\begin{equation}
\mathbf{x} = [\boldsymbol{\theta}_1^{\Ttran}, \boldsymbol{\theta}_2^{\Ttran}, \ldots, \boldsymbol{\theta}_K^{\Ttran}]^{\Ttran}.
\end{equation}
This structure is referred to as an \textit{expanded representation} because the dimensionality of each individual grows proportionally with the number of sources, thereby expanding the search space to jointly cover all source parameters within a single optimization vector. 
Such a formulation enables the evolutionary process to estimate all source locations simultaneously rather than sequentially. 
%As a result, NEEF-DE mitigates error accumulation across detection stages,
In contrast to NEMO-DE, which explicitly enforces multimodality in the search process, NEEF-DE operates within a unified parameter space where the optimal solution corresponds to the joint configuration that best reproduces the received signal subspace.

\subsubsection{Eigen-Subspace Fitting Objective Function}
For expanded representation, we define an objective function based on subspace misalignment, inspired by subspace-fitting methods such as Weighted Subspace Fitting (WSF)~\cite{viberg1991wsf}, which quantifies the mismatch between the received signal subspace and the model subspace spanned by the candidate array response matrix.

Similar to subspace-based localization methods such as MUSIC, we first form the sample covariance matrix $\mathbf{R}$ and extract its top $K$ eigenvectors to obtain the received signal subspace. Specifically, the sample covariance matrix is computed as 
\begin{equation}
 \mathbf{R} = \frac{1}{T} \mathbf{Y}\mathbf{Y}^{\Htran}, 
\end{equation}
and its eigendecomposition yields
\begin{equation}
   \mathbf{R} = \mathbf{U}_s\boldsymbol{\Sigma}_s \mathbf{U}_s^{\Htran} +   \mathbf{U}_n\boldsymbol{\Sigma}_n \mathbf{U}_n^{\Htran}, 
\end{equation}
where $\boldsymbol{\Sigma}_s \in \mathbb{C}^{K\times K}$ and $\boldsymbol{\Sigma}_n \in \mathbb{C}^{(M-K)\times (M-K)}$ contain the $K$ largest and $(M-K)$ smallest eigenvalues of $\mathbf{R}$ on their diagonal, and $\mathbf{U}_s \in \mathbb{C}^{M \times K}$ and $\mathbf{U}_n \in \mathbb{C}^{M \times (M-K)}$ consist of the corresponding eigenvectors.  
The eigen-subspace fitting (ESF) objective function is defined as 
\begin{equation}
  J_{\mathrm{ESF}}(\mathbf{x}) = \|(\mathbf{I} - \mathbf{P}_{\mathbf{A}}(\mathbf{x}))\mathbf{U}_s\|_F^2,  
\end{equation}
where $\mathbf{P}_{\mathbf{A}}(\mathbf{x}) = \mathbf{A} \left( \mathbf{A}^{\Htran} \mathbf{A}\right)^{-1} \mathbf{A}^{\Htran}$ and $\mathbf{A}$ is defined after \eqref{eq:collected_signal}.

\begin{algorithm}[t]
\caption{Joint ESF-based DE Localization}
\label{Alg:NEEF}
\KwInput{Received data $\mathbf{Y}$, number of sources $K$, joint search domain $\boldsymbol{\Theta}_{\mathrm{joint}} \subseteq \mathbb{R}^{2K}$, population size $N_p$, max generations $G_{\max}$, DE parameters $(F,C_r)$}
\KwOutput{Joint estimate $\hat{\mathbf{x}} = [\hat{\boldsymbol{\theta}}_1^{\Ttran}, \dots, \hat{\boldsymbol{\theta}}_K^{\Ttran}]^{\Ttran}$}
\BlankLine

\tcp{Compute signal subspace once}
$T \leftarrow$ number of snapshots\;
$\mathbf{R} \leftarrow \frac{1}{T}\mathbf{Y}\mathbf{Y}^{\Htran}$\;
Compute eigendecomposition $\mathbf{R} = \mathbf{U}\boldsymbol{\Sigma}\mathbf{U}^{\Htran}$\;
Construct $\mathbf{U}_s$ from the $K$ dominant eigenvectors\;

\BlankLine
\tcp{Define ESF objective function}
\For{candidate joint vector $\mathbf{x} = [\boldsymbol{\theta}_1^{\Ttran}, \dots, \boldsymbol{\theta}_K^{\Ttran}]^{\Ttran}$}{
  Construct array response matrix $\mathbf{A}(\mathbf{x}) = [\mathbf{a}(\boldsymbol{\theta}_1), \dots, \mathbf{a}(\boldsymbol{\theta}_K)]$\;
  Compute projection matrix $\mathbf{P}_{\mathbf{A}}(\mathbf{x}) = 
    \mathbf{A}(\mathbf{x})(\mathbf{A}^{\Htran}(\mathbf{x})\mathbf{A}(\mathbf{x}))^{-1}\mathbf{A}^{\Htran}(\mathbf{x})$\;
  $J_{\mathrm{ESF}}(\mathbf{x}) \leftarrow \|(\mathbf{I} - \mathbf{P}_{\mathbf{A}}(\mathbf{x}))\mathbf{U}_s\|_F^2$\;
}

\BlankLine
\tcp{Initialize population}
Generate $N_p$ random candidates $\{\mathbf{x}^{(1)}, \ldots, \mathbf{x}^{(N_p)}\}$ within $\boldsymbol{\Theta}_{\mathrm{joint}}$\;
Evaluate $J_{\mathrm{ESF}}(\mathbf{x}^{(i)})$ for all individuals\;

\BlankLine
\tcp{Run DE (or any evolutionary optimizer) in the joint $2K$-dimensional space}
Run DE for $G_{\max}$ generations to minimize $J_{\mathrm{ESF}}(\mathbf{x})$\;

\BlankLine
\tcp{Return best estimate}
Select $\hat{\mathbf{x}} = \arg\min_{\mathbf{x}^{(i)}} J_{\mathrm{ESF}}(\mathbf{x}^{(i)})$\;

\Return $\hat{\mathbf{x}}$
\end{algorithm}

\subsubsection{Joint Evolutionary Search Strategy}
Given the expanded representation and the ESF objective function, NEEF-DE employs a joint evolutionary search to estimate all source parameters simultaneously. %Unlike the proposed multimodal framework, NEMO-DE, which detect sources sequentially, NEEF-DE evolves a population of complete multi-source hypotheses within a single optimization process.
It is worthwhile to mention that the term \textit{joint} reflects the fact that NEEF-DE estimates the parameters of all $K$ sources within a single optimization vector and evolves a population of complete multi-source hypotheses within a single optimization process. Instead of detecting sources one-by-one, each individual in the population encodes the complete set of location parameters, and the cost function evaluates the collective consistency of all hypothesized sources with the received signal subspace. As a result, the algorithm optimizes all $2K$ parameters simultaneously, allowing the evolutionary process to account for spatial interactions and joint subspace structure induced by the multiple near-field sources. This simultaneous, all-at-once estimation is the primary reason the proposed method is referred to as a \textit{joint} evolutionary search.

\subsection{Joint Evolutionary Process}
At the beginning of the search, a population 
\(\{\mathbf{x}^{(1)}, \ldots, \mathbf{x}^{(N_p)}\}\) 
is initialized by sampling each parameter uniformly within its feasible bounds. 
Each population member therefore corresponds to a full joint hypothesis of all $K$ source locations, providing a diverse coverage of the $2K$-dimensional search space.

%\textcolor{red}{The proposed joint representation and objective formulation are independent of the specific evolutionary search strategy and can be integrated with a wide range of population-based optimization methods. In this work, DE is adopted as a representative solver.} 
The evolutionary process follows the standard DE/rand/1/bin scheme (Section~\ref{sec:DE}).
 For each target vector \(\mathbf{x}^{(i)}\), three distinct individuals are selected to construct a mutant vector, which is then combined with the target through binomial crossover to form a trial vector. A deterministic selection step is applied: the trial vector replaces the target only if it yields a lower ESF cost value. This cost-driven replacement gradually guides the population toward configurations whose array response matrices exhibit stronger alignment with the observed signal subspace.

\begin{table*}[t]
\centering
\caption{Comparison Between NEMO-DE and NEEF-DE Localization Frameworks}
\label{tab:comparison}
\renewcommand{\arraystretch}{1.25}
\begin{tabular}{p{3.2cm} p{5.2cm} p{5.2cm}}
\toprule
\textbf{Aspect} & \textbf{NEMO-DE (Sequential Multimodal DE)} & \textbf{NEEF-DE (Joint Subspace Fitting DE)} \\
\midrule

\textbf{Representation} &
Single-source representation $\boldsymbol{\theta} = [\phi, r]^{\Ttran}$; each DE run estimates one source location. &
Expanded joint representation $\mathbf{x} = [\boldsymbol{\theta}_1^{\Ttran},\dots,\boldsymbol{\theta}_K^{\Ttran}]^{\Ttran}$; all source locations are estimated simultaneously. \\

\textbf{Search Strategy} &
Sequential multimodal evolutionary search; $K$ DE runs. &
Single joint evolutionary search in a $2K$-dimensional space; one DE run. \\

\textbf{Objective Function} &
Residual reconstruction error in the data domain; depends on residual signal after each detection. &
Subspace mismatch; uses the signal subspace and is evaluated once per candidate solution. \\

\textbf{Multimodality Handling} &
Explicit multimodality via repeated searches, penalization, and residual updates. &
Implicit multimodality arising from joint parameterization and the objective function landscape. \\

\textbf{Residual Update} &
Yes; contribution of each detected source is subtracted after each run. &
No residual update required; all sources modeled jointly. \\

\textbf{Penalization} &
Explicit distance-based penalization to avoid redundant detections. &
Not required; joint search inherently avoids repeated solutions. \\

\textbf{Number of Optimizations} &
$K$ optimization runs (one per detected source). &
Exactly one optimization run regardless of $K$. \\

\textbf{Error Propagation} &
Possible, since each detection affects the next residual. &
Avoided; all sources are estimated in a unified step. \\

\bottomrule
\end{tabular}
\end{table*}

Since optimization occurs directly in the joint parameter space, NEEF-DE does not require explicit mechanisms for maintaining diversity, enforcing mode separation, or updating residuals. Multimodality is resolved implicitly through the optimization landscape, whose global minimizer corresponds to the full multi-source configuration that best fits the observed data. The evolutionary process continues until a predefined number of generations is reached or until the improvement in the best objective value falls below a convergence threshold. The best individual in the final population is then taken as the joint estimate of all source locations. The pseudocode of the NEEF-DE scheme is provided in  Algorithm~\ref{Alg:NEEF}.

For improved clarity, Table~\ref{tab:comparison} highlights the key differences between the two proposed approaches.

\section{Computational Complexity Analysis}
\label{sec:complexity}
\textbf{NEMO-DE:} For each candidate $\boldsymbol{\theta}$, evaluating $J_{\mathrm{RLS}}(\boldsymbol{\theta})$ in \eqref{eq:RLS} is dominated by forming the projection of $\mathbf{Y}$ onto $\mathbf{a}(\boldsymbol{\theta})$ which scales as $O(MT)$. Computing the Frobenius norm of the resulting residual matrix also scales as $O(MT)$. Hence, the per-candidate fitness evaluation has complexity $O(MT)$.
NEMO-DE evaluates approximately $N_p$ candidate solutions per generation over $G_{\mathrm{max}}$ generations. Therefore, the computational complexity for estimating the location of a single source 
is $O(N_p G_{\mathrm{max}} MT)$. Since NEMO-DE localizes the sources sequentially, repeating this procedure $K$ times results in the overall complexity of $O(K N_p G_{\mathrm{max}} MT)$.

\textbf{NEEF-DE:} Forming the sample covariance matrix $\mathbf{R}$ costs $O(M^2 T)$ and performing the eigenvalue decomposition of the resulting matrix scales as $O(M^3)$. For each candidate solution, the complexity is dominated by computing the Gram matrix $\mathbf{A}^{\Htran} \mathbf{A}$ and applying the projector $\mathbf{P}_{\mathbf{A}}$ to $\mathbf{U}_s$, both scaling as $O(MK^2)$. During the DE fitness evaluations, approximately $N_p$ candidates are evaluated per generation over $G_{\mathrm{max}}$ generations. Therefore, the overall computational complexity for NEEF-DE is $O(M^2T + M^3 + N_p G_{\mathrm{max}} M K^2)$.

\textbf{2D MUSIC:} Similar to NEEF-DE, the formation of the sample covariance and performing the eigenvalue decomposition incur the complexity of $O(M^2 T)$ and $O(M^3)$, respectively. MUSIC evaluates the pseudospectrum by computing the projection $\mathbf{U}_n^{\Htran}\, \mathbf{a}(\phi,r)$ whose complexity is $O(M(M-K)) \approx O(M^2)$. With $G_{\mathrm{angle}}$ angle points and $G_{\mathrm{range}}$ range points, the total grid search complexity scales as $O(G_\phi G_r M^2)$, yielding an overall complexity of $O(M^2 T + M^3 + G_\phi G_r M^2)$ for 2D MUSIC. 

When extending the above frameworks to 3D localization by including the elevation angle, the complexities of NEMO-DE and NEEF-DE remain unchanged if $N_p$ and $G_{\max}$ are kept the same, since the complexity of dominant operations in each fitness evaluation is determined by $M$, $T$, and $K$. In contrast, MUSIC requires a 3D grid search, introducing an elevation grid with $G_\psi$ points. Hence, the complexity of the grid search increases to $O(G_\phi G_\psi G_r M^2)$. Since MUSIC typically requires a fine grid to achieve satisfactory localization accuracy, its computational cost becomes prohibitive in 3D localization scenarios.

\section{Performance Evaluation}
\label{sec:evaluation}
Here, we evaluate the performance of the proposed frameworks. We first compare NEMO-DE  with two well-known high-resolution localization methods: 2D MUSIC \cite{Huang1991Near} and its modified low-complexity version \cite{He2012Efficient}. We first describe the simulation setup and then present extensive numerical evaluations of the proposed methods against benchmarks. We then discuss the main limitations of the proposed frameworks, thereby identifying directions for future research in this area.

\subsection{Simulation Setup}
The channels are modeled by Rician fading as
\begin{equation}
\mathbf{h}_k = \sqrt{\beta_k}\left(\sqrt{\frac{\kappa}{\kappa + 1}} \mathbf{h}_{k,\mathrm{LoS}} + \sqrt{\frac{1}{\kappa + 1}} \mathbf{h}_{k,\mathrm{NLoS}}\right),
\end{equation}
where $\beta_k$ is the path-loss for source~$k$, $\kappa$ denotes the Rician factor, $\mathbf{h}_{k,\mathrm{LoS}}$ is the LoS component of the channel characterized by the array response vector, and $\mathbf{h}_{k,\mathrm{NLoS}}$ is the NLoS component consisting of scattered multi-path channels based on correlated Rayleigh fading, as described in \cite{Demir2024Spatial}. 
The following parameters are used throughout the simulations: The Rician factor is set as $\kappa = 10$, the wavelength equals to $\lambda = 0.02\,$m corresponding to the carrier frequency of $15\,$GHz, and the number of samples is set to be $T = 200$. The large-scale path-loss factor $\beta_k,\,\forall k$ is absorbed into the specified SNR values.
The values of other parameters used in Algorithm~\ref{Alg:NEMO} are given in Table~\ref{tab:DE_params}. The parameters are selected based on a preliminary sensitivity analysis, in which multiple parameter configurations were systematically evaluated. 
\vspace{-1mm}
\begin{table}[H]
  \centering
  \caption{DE parameter values.}
  \label{tab:DE_params}
  \normalsize          % <-- add this line (or \large)
  \begin{tabular}{lc}
    \toprule
    \textbf{Parameter} & \textbf{Value} \\
    \midrule
    $N_p$              & 50    \\
    $G_{\max}$ & 300   \\
    $F$                & 0.5   \\
    $C_r$              & 0.8   \\
    $\delta_{\mathrm{min}}$ & 0.08 \\
    $\alpha$           & 1000  \\
    \bottomrule
  \end{tabular}
  \vspace{-2mm}
\end{table}

For performance evaluation, we use the root mean square error (RMSE) computed in Cartesian coordinates. In specific, the estimated angle and range parameters are first converted to Cartesian coordinates, and the RMSE is then calculated between the true and estimated source positions.

We consider a ULA of $M = 128$ antennas at the BS, and $K = 3$ sources to be localized. The spacing between the antennas is assumed to be $\lambda/4$. The sources' azimuth angles and ranges are randomly selected from uniform distributions $\mathcal{U}[-60^\circ,60^\circ]$ and $\mathcal{U}[2D_{\mathrm{ap}},d_{\mathrm{FA}}/2]$, respectively. In the considered setup, $D_{\mathrm{ap}} = 0.635\,$m and $d_{\mathrm{FA}} \approx 40.3\,$m.
For MUSIC and modified MUSIC, we use two grid sizes. One with $G_{\mathrm{angle}} = 200$ grid points for the angle and $G_{\mathrm{range}} = 1000$ grid points for the range, corresponding to step sizes of $\Delta \phi = 0.6^\circ$ and $\Delta r \approx 1.9\,$cm, respectively, and another with double the grid size of the first one, resulting in $\Delta \phi = 0.3^\circ$ and $\Delta r \approx 0.95\,$cm. 

\begin{figure}
    \centering
    \includegraphics[width=0.9\linewidth]{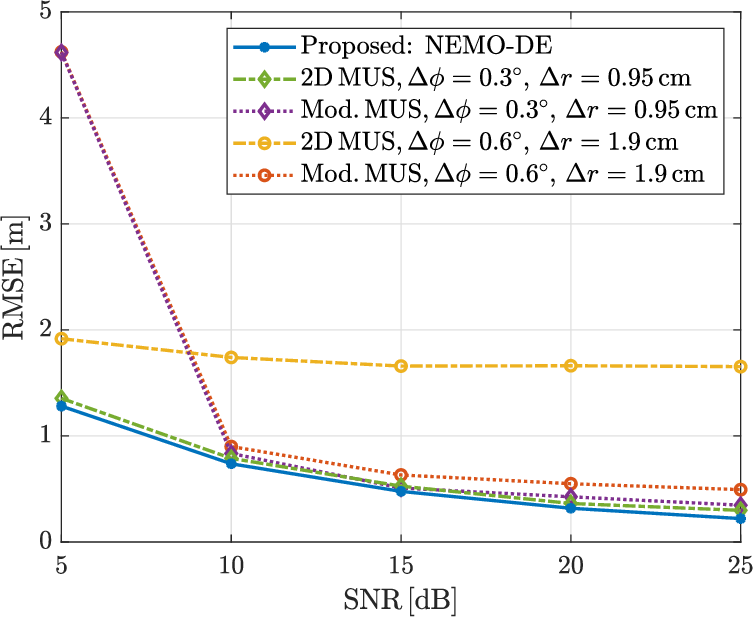}
    \caption{RMSE vs. SNR for NEMO-DE, 2D MUSIC, and modified MUSIC. The BS antennas are spaced by $\lambda/4$.}
\label{fig:rmse_snr1}
\end{figure}

\begin{figure}
    \centering
    \includegraphics[width=0.9\linewidth]{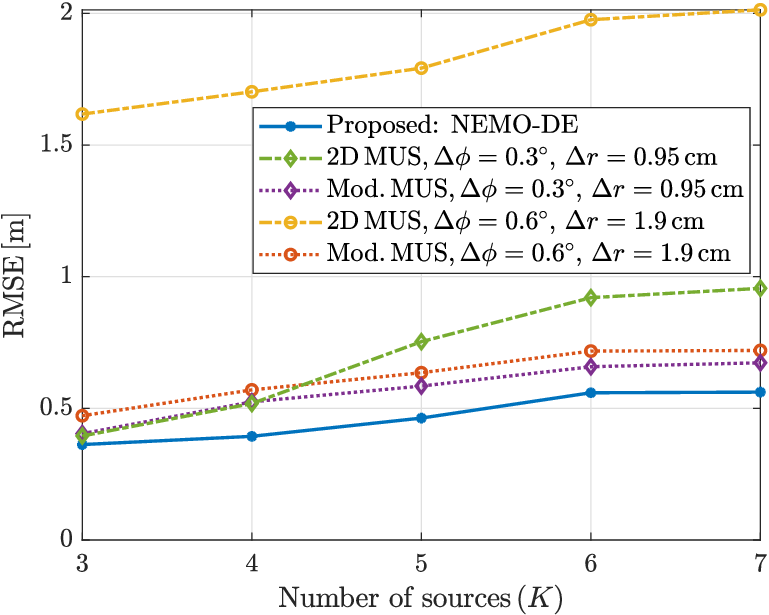}
    \caption{RMSE vs. number of sources for NEMO-DE, 2D MUSIC, and modified MUSIC. The BS antennas are spaced by $\lambda/4$.}
\label{fig:rmse_K1}
\end{figure}

\begin{figure}
    \centering
    \includegraphics[width=0.9\linewidth]{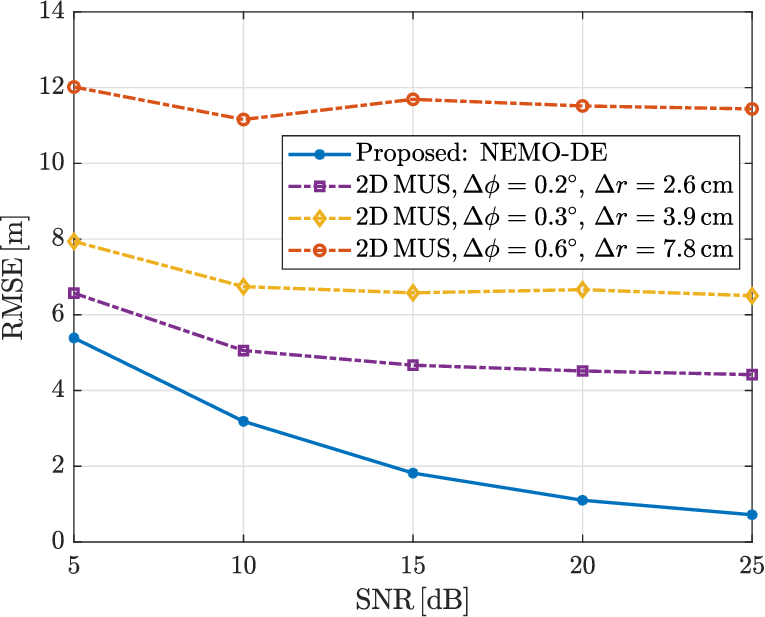}
    \caption{RMSE vs. SNR for NEMO-DE and 2D MUSIC. The BS antennas are spaced by $\lambda/2$.}
\label{fig:rmse_snr2}
\end{figure}

\begin{figure}
    \centering
    \includegraphics[width=0.9\linewidth]{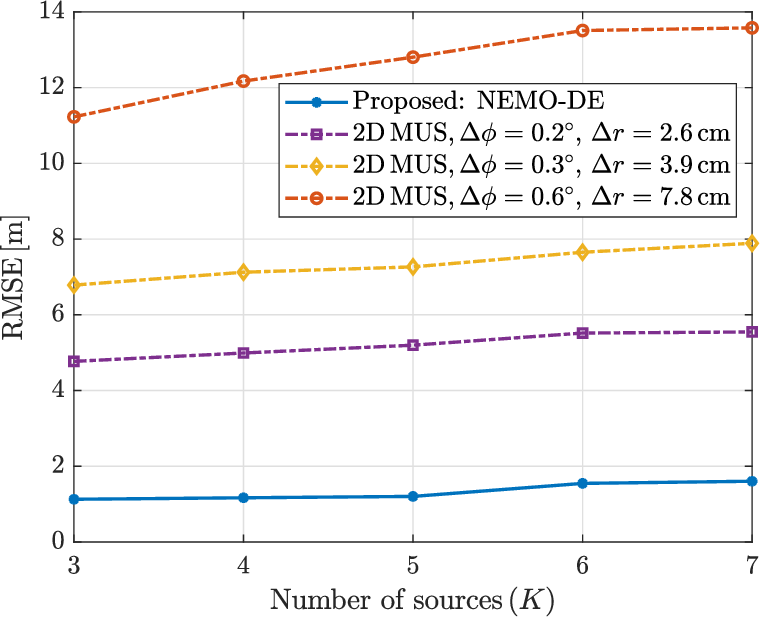}
    \caption{RMSE vs. number of sources for NEMO-DE and 2D MUSIC. The BS antennas are spaced by $\lambda/2$.}
\label{fig:rmse_K2}
\end{figure}

\subsection{Results}
Fig.~\ref{fig:rmse_snr1} depicts the performance of the proposed NEMO-DE, 2D MUSIC, and modified MUSIC as a function of SNR. We can  see that the proposed method exhibits a promising RMSE across all considered SNR values, and, in contrast to the MUSIC and modified MUSIC, it does not rely on any angle-range grid. 
We also observe that 2D MUSIC is more sensitive to grid mismatch than modified MUSIC because it performs a joint 2D search over angle and range, while the modified MUSIC decouples angle and range estimation into successive 1D searches and is therefore less affected by discretization.

In Fig.~\ref{fig:rmse_K1}, we investigate the performance of NEMO-DE, 2D MUSIC, and modified MUSIC for different numbers of sources. The same setup as in the previous simulation is used here and  SNR equals $20\,$dB for all sources. The RMSE gradually increases with the number of sources. In the proposed method, each evaluation of the objective function $J_{\mathrm{RLS}}(\boldsymbol{\theta})$ for one source treats all other sources as structured interference embedded in $\mathbf{Y}$, making the localization problem more challenging and leading to increased RMSE as the number of sources increases. Moreover, for MUSIC-based methods, increasing the number of sources reduces the separation between signal and noise subspaces, again leading to increased average localization error. 
\begin{figure}[!t]
  \centering

  \begin{subfigure}{\columnwidth}
    \centering
    \includegraphics[width=0.9\linewidth]{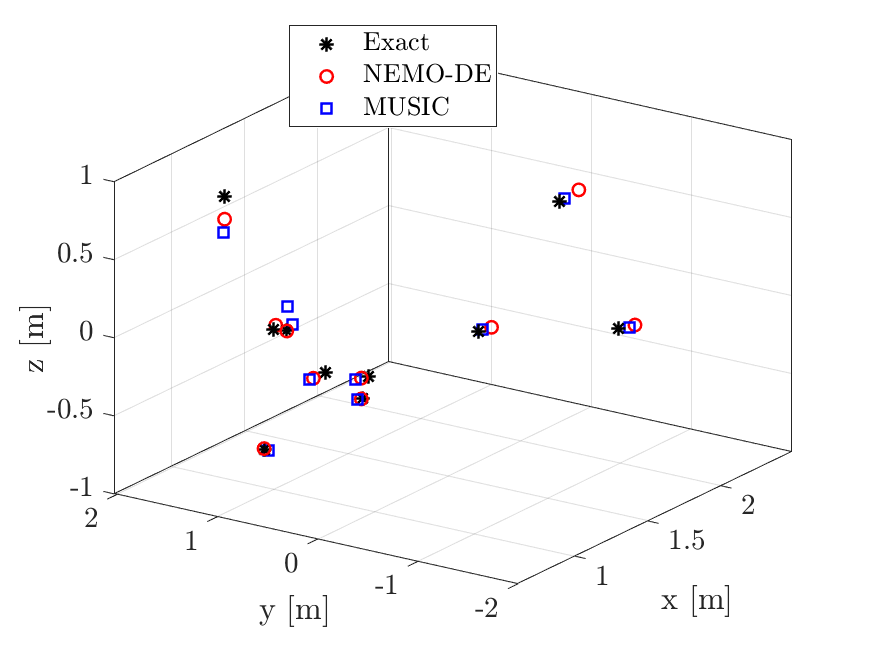}
    \caption{Source 1}
    \label{fig:source1_NEMO}
  \end{subfigure}

 % \vspace{0.6em}

  \begin{subfigure}{\columnwidth}
    \centering
    \includegraphics[width=0.9\linewidth]{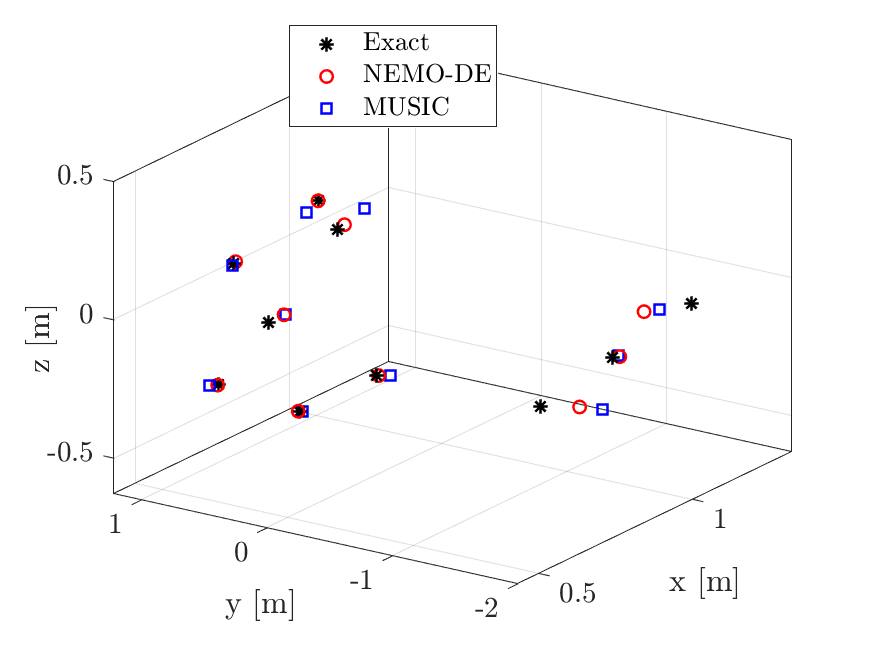}
    \caption{Source 2}
    \label{fig:source2_NEMO}
  \end{subfigure}

  %\vspace{0.6em}

  \begin{subfigure}{\columnwidth}
    \centering
    \includegraphics[width=0.9\linewidth]{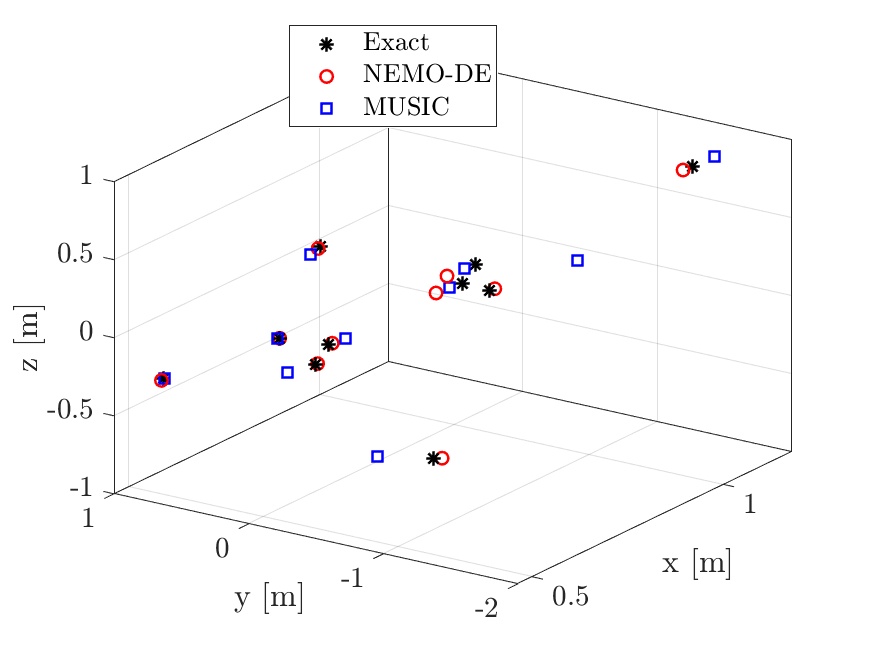}
    \caption{Source 3}
    \label{fig:source3_NEMO}
  \end{subfigure}

  \caption{Exact locations and estimated ones by NEMO-DE and MUSIC for three sources over $10$ random realizations.}
  \label{fig:locations_NEMO}
\end{figure}

\begin{figure}
    \centering
    \includegraphics[width=0.9\linewidth]{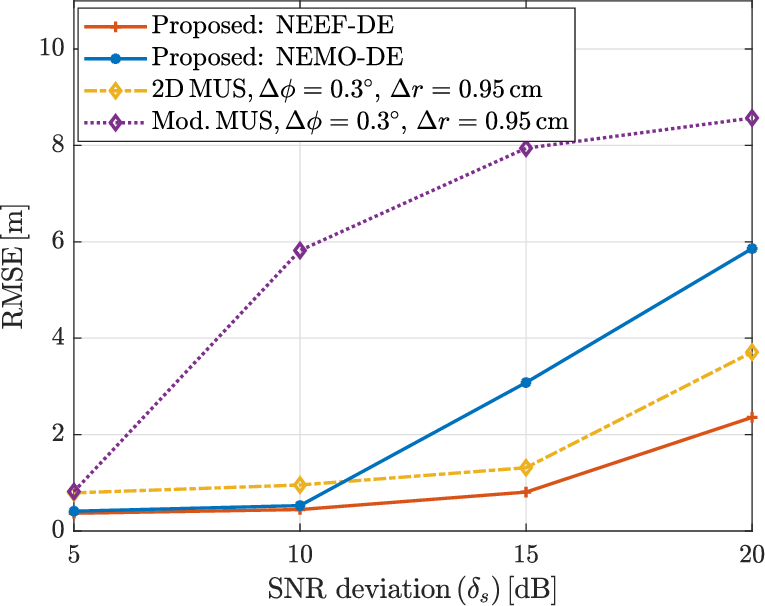}
    \caption{RMSE of NEEF-DE, NEMO-DE, 2D MUSIC, and modified MUSIC under unequal received SNRs across sources.}
\label{fig:rmse_snrdev2}
\vspace{-3mm}
\end{figure}

\begin{figure}[!t]
  \centering

  \begin{subfigure}{\columnwidth}
    \centering
    \includegraphics[width=0.9\linewidth]{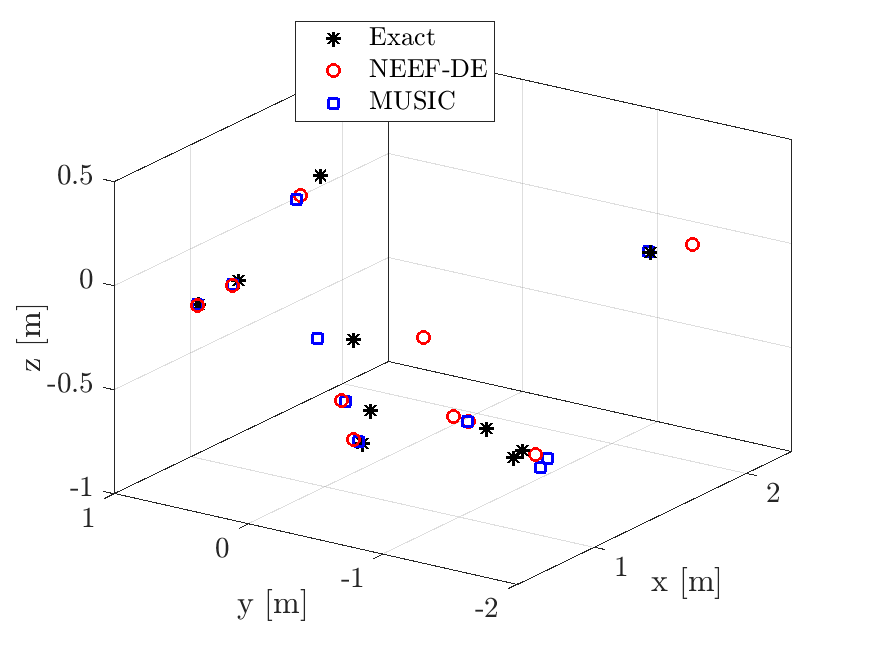}
    \caption{Source 1}
    \label{fig:source1_NEEF}
  \end{subfigure}

 % \vspace{0.6em}

  \begin{subfigure}{\columnwidth}
    \centering
    \includegraphics[width=0.9\linewidth]{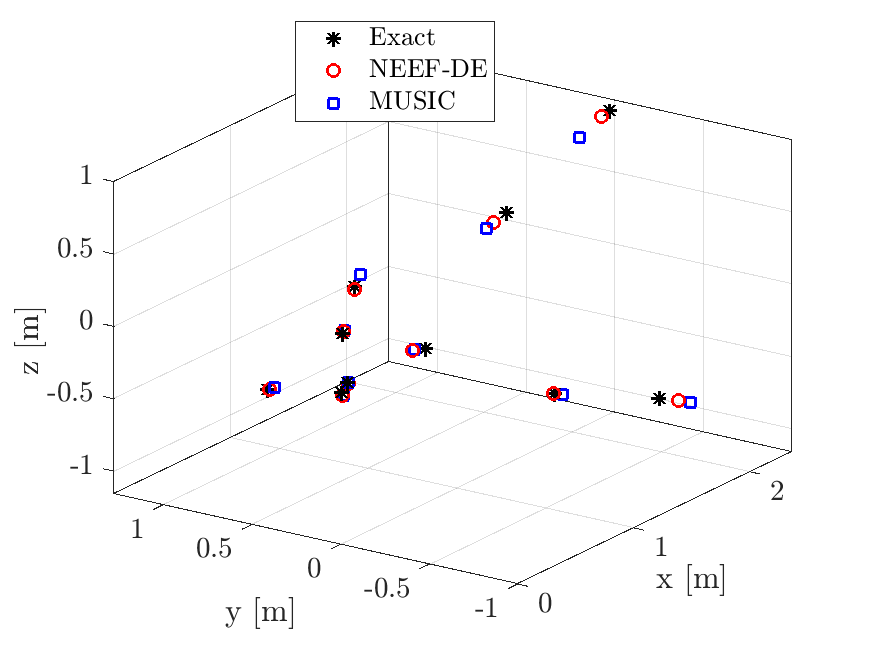}
    \caption{Source 2}
    \label{fig:source2_NEEF}
  \end{subfigure}

 % \vspace{0.6em}

  \begin{subfigure}{\columnwidth}
    \centering
    \includegraphics[width=0.9\linewidth]{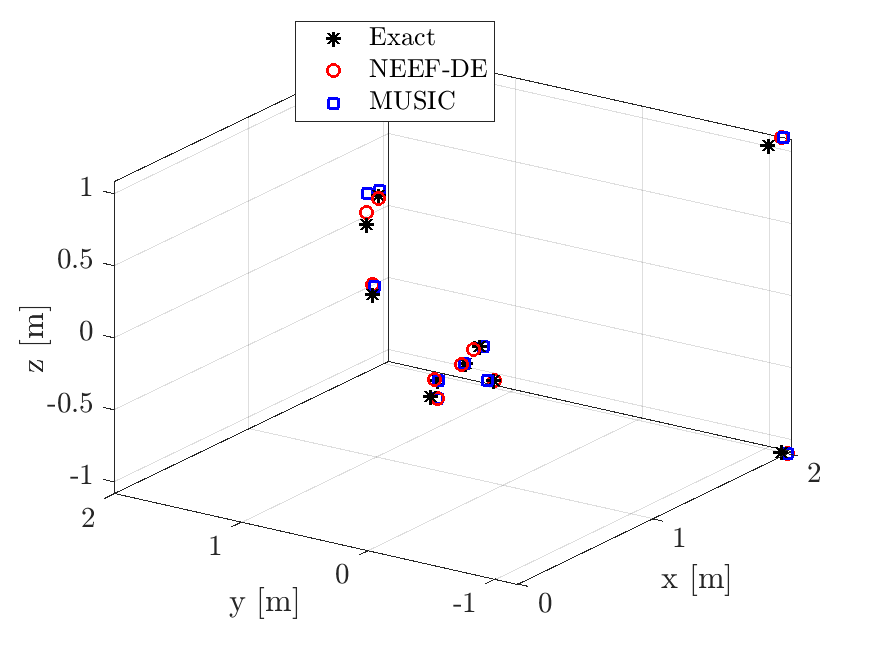}
    \caption{Source 3}
    \label{fig:source3_NEEF}
  \end{subfigure}

  \caption{Exact locations and estimated ones by NEEF-DE and MUSIC for three sources over $10$ random realizations.}
  \label{fig:locations_NEEF}
\end{figure}

Next, we evaluate the performance of NEMO-DE and 2D MUSIC for the case where the antenna spacing in the ULA is $\lambda/2$. The modified MUSIC cannot be evaluated in this case since it requires a spacing of less than $\lambda/4$, a major limitation of this method which hinders its practical implementation. In this case, we have $D_{\mathrm{ap}} = 1.27\,$m and $d_{\mathrm{FA}} \approx 161\,$m. 2D MUSIC is evaluated for three grid sizes: one with $G_{\mathrm{angle}} = 200$ and $G_{\mathrm{range}} = 1000$, corresponding to $\Delta \phi = 0.6^\circ$ and $\Delta r \approx 7.8\,$cm, one with $G_{\mathrm{angle}} = 400$ and $G_{\mathrm{range}} = 2000$, yielding $\Delta \phi = 0.3^\circ$ and $\Delta r \approx 3.9\,$cm, and one with $G_{\mathrm{angle}} = 600$ and $G_{\mathrm{range}} = 3000$, resulting in $\Delta \phi = 0.15^\circ$ and $\Delta r \approx 2.6\,$cm. The performance as a function of SNR and number of sources is depicted in Fig.~\ref{fig:rmse_snr2} and Fig.~\ref{fig:rmse_K2}, respectively. Similar observations to those in Fig.~\ref{fig:rmse_snr1}
and Fig.~\ref{fig:rmse_K1} can be made here. We also observe that the 2D MUSIC performs poorly even with a relatively fine grid. We can keep increasing the grid resolution for MUSIC until we achieve a satisfactory performance. However, the complexity of 2D MUSIC grows quadratically with the grid resolution, making it inefficient for precise localization. This issue is exacerbated in 3D localization, where the BS is in the form of a UPA and the location parameter vector $\boldsymbol{\theta} = [\phi,\psi,r]^{\Ttran}$ has three parameters.

We next consider a UPA-formed BS and compare the 3D localization performance of NEMO-DE and 3D MUSIC. Fig.~\ref{fig:locations_NEMO} visualizes the exact and estimated 3D locations of $K = 3$ sources over $10$ random realizations.
The setup includes a $16\times 16$ UPA at the BS with $\lambda/2$ inter-antenna spacing in both dimensions, $T = 100$ samples, and SNR of $20\,$dB. The array aperture and Fraunhofer distance for this setup are $D_{\mathrm{ap}} = 0.226\,$m and $d_{\mathrm{FA}} = 5.12\,$m. The sources' azimuth and elevation angles are randomly selected from $\mathcal{U}[-60^\circ,60^\circ]$ and $\mathcal{U}[-30^\circ,30^\circ]$, respectively, while the ranges are picked from $\mathcal{U}[2D_{\mathrm{ap}},d_{\mathrm{FA}}/2]$. 
For 3D MUSIC, $G_{\mathrm{angle}} = 200$ and $G_{\mathrm{range}} = 400$ are considered, resulting in angular and range resolutions of $\Delta \phi = 0.6^\circ$, $\Delta \psi = 0.3^\circ$, and $\Delta r = 0.5\,$cm.  In Fig.~\ref{fig:locations_NEMO}, black stars indicate the exact locations, red circles represent the estimated locations via NEMO-DE, and blue squares show the estimated locations by MUSIC. 
Overall, NEMO-DE attains accuracy comparable to 3D MUSIC, and in several realizations yields estimates closer to the exact locations, while avoiding MUSIC’s computationally expensive 3D grid search and the associated complexity burden. 

We now evaluate the performance of the proposed NEEF-DE framework under SNR imbalance among sources. We consider the same setup used in Fig.~\ref{fig:rmse_snr1} with $\Delta \phi = 0.3^\circ$ and $\Delta r = 0.95\,$cm for 2D MUSIC and modified MUSIC. We fix the SNR of one source at $20\,$dB and set the SNRs of the other two sources to $20-\delta_s\,$dB and $20+\delta_s\,$dB, where $\delta_s$ is varied from $5\,$dB to $20\,$dB in steps of $5\,$dB. For NEEF-DE, we use $N_p = 40K$ and $G_{\max} = 500$, while $F$ and $C_r$ are set according to Table~\ref{tab:DE_params}.
The results are shown in Fig.~\ref{fig:rmse_snrdev2}. As $\delta_s$ increases, the NEEF-DE scheme exhibits a much more stable RMSE than NEMO-DE, showing only mild variations across the considered SNR deviations, whereas the performance of NEMO-DE degrades more noticeably with increasing SNR imbalance. The reason is that with NEMO-DE, sources with higher SNR dominate the residual and distort the cost landscape when the SNRs are highly unbalanced. 
In contrast, NEEF-DE relies on a subspace fitting criterion that matches the model-based and data-derived signal subspaces, which depend primarily on the array response vectors rather than the exact source powers; therefore, NEEF-DE remains much more robust to SNR variations. We also observe that the modified MUSIC scheme performs poorly under SNR imbalance. That's because in the modified MUSIC algorithm, the covariance matrix is first compressed into an anti-diagonal sequence before angle estimation. This sequence is essentially a power-weighted mixture of all sources, so a strong source dominates the resulting structure.

Fig.~\ref{fig:locations_NEEF} compares the 3D location estimation performance of NEEF-DE and 3D MUSIC over $10$ random location realizations.
The setup follows that of Fig.~\ref{fig:locations_NEMO}, except that the three sources’ SNRs are set to $10$, $20$, and $30\,$dB. We observe the same trend as in Fig.~\ref{fig:locations_NEMO}: our proposed DE-based scheme maintains comparable accuracy to 3D MUSIC with substantially lower complexity.
\begin{comment}
\begin{figure*}[t]
	\centering
	\begin{subfigure}{0.325\textwidth}
		\centering
		\includegraphics[width=\linewidth]{visualization_NEEFvsMUSIC_S1.eps}
		\caption{Source 1}
		\label{fig:source1_NEEF}
	\end{subfigure}
	\hfill
	\begin{subfigure}{0.325\textwidth}
		\centering
		\includegraphics[width=\linewidth]{visualization_NEEFvsMUSIC_S2.eps}
		\caption{Source 2}
		\label{fig:source2_NEEF}
	\end{subfigure}
	\hfill
	\begin{subfigure}{0.325\textwidth}
		\centering
		\includegraphics[width=\linewidth]{visualization_NEEFvsMUSIC_S3.eps}
		\caption{Source 3}
		\label{fig:source3_NEEF}
	\end{subfigure}
	\caption{Exact locations and estimated ones by NEEF-DE and MUSIC for three sources over $10$ random realizations.}
	\label{fig:locations_NEEF}
\end{figure*}
\end{comment}

To provide insight into the complexity-performance trade-off between 3D MUSIC and the proposed DE-based localization schemes, Table~\ref{tab:runtime-rmse} shows the runtime and RMSE values of 3D localization for a single realization of NEMO-DE, NEEF-DE, and MUSIC under the setup considered in Fig.~\ref{fig:locations_NEMO} and Fig.~\ref{fig:locations_NEEF}, and the SNR is set as $20\,$dB for all sources. The results have been obtained using MATLAB on a computer with an Intel Core i5-1145G7 CPU @\,2.60\,GHz, 16\,GB RAM, running a 64-bit Windows operating system. NEMO-DE exhibits the lowest runtime, which is consistent with the computational complexity analysis in Section~\ref{sec:complexity}. NEEF-DE incurs a higher runtime than NEMO-DE, yet it is substantially faster than 3D MUSIC. Note that the key advantage of NEEF-DE over NEMO-DE emerges under SNR imbalance across sources. When the sources have nearly equal SNRs, NEMO-DE is generally preferable due to its lower computational cost.

\begin{table}[h]
\centering
\caption{Runtime and RMSE for DE-based schemes and MUSIC.}
\renewcommand{\arraystretch}{1.2} % increase row height a bit
\begin{tabular}{|
  >{\centering\arraybackslash}p{0.45\linewidth} |
  >{\centering\arraybackslash}p{0.20\linewidth} |
  >{\centering\arraybackslash}p{0.20\linewidth} |
}
\hline 
Method              & Runtime [s] & RMSE [m] \\ \specialrule{.25em}{.07em}{.075em}
NEMO-DE             & 4.101       & 0.0293   \\ \hline
NEEF-DE  & 70.64       & 0.0788    \\ \hline
3D MUSIC  & 355.3       & 0.0687   \\ \hline
\end{tabular}
\label{tab:runtime-rmse}
\end{table}
%Jalal: ablation study
\subsection{Limitations}
The proposed frameworks provide grid-free continuous-domain alternatives to MUSIC-type near-field localization methods, but they also have several limitations. First, both NEMO-DE and NEEF-DE rely on the existence of a sufficiently dominant source-dependent near-field array response. Therefore, the present formulation is most suitable for LoS-dominant or mildly scattered Rician propagation conditions, and the results should not be interpreted as demonstrating robustness to severe NLoS scenarios. Second, the number of sources is assumed to be known or estimated before localization. Although information-theoretic criteria such as AIC and MDL can be used for source enumeration, source-number mismatch may affect the two frameworks differently. In particular, NEEF-DE is more sensitive to an incorrect value of $K$ because its representation dimension depends directly on the assumed number of sources. Third, although the proposed methods avoid  angle-range grid construction, they still require stochastic numerical optimization. Their performance and runtime therefore depend on evolutionary search parameters such as population size, number of generations, mutation factor, crossover rate, initialization bounds, and stopping criteria. Finally, NEEF-DE improves robustness to source-power imbalance by estimating all sources jointly, but this comes at the cost of higher computational burden compared with NEMO-DE. These observations indicate that the proposed methods should be viewed as flexible grid-free alternatives to MUSIC-type searches. A systematic study of source enumeration, source-number mismatch, optimizer selection, and adaptive parameter tuning is therefore an important direction for future work.

\section{Conclusion}
\label{sec:conc}
This work introduced evolutionary optimization as a continuous-domain search tool for near-field multi-source localization.  We developed two complementary evolutionary frameworks based on DE as a representative algorithm, one built on residual data fitting and one on signal subspace fitting, that overcome key limitations of grid-based MUSIC-type methods as well as the generalization challenges associated with data-dependent deep learning approaches.  Specifically, the first method, NEMO-DE, adopts a compact single-source representation and applies multimodal DE over a residual reconstruction error objective.
It performs sequential searches with projection-based residual updates and distance-based penalization to estimate multiple near-field source locations.
Our simulations showed that NEMO-DE achieves competitive localization accuracy relative to MUSIC and its modified low-complexity variant, while working in the continuous parameter space and avoiding any grid discretization. 
However, the performance of NEMO-DE degrades when the sources exhibit large SNR disparities, since the energy-weighted residual fitting tends to be dominated by strong sources. Motivated by this, we proposed the second framework, NEEF-DE, which encodes all sources jointly and minimizes a subspace-misalignment objective.
Numerical simulations demonstrate that this method maintains stable localization accuracy across pronounced SNR variations.

\bibliographystyle{IEEEtran}
\bibliography{literature}
\end{document}